\documentclass[journal]{journal}
\usepackage{glossaries}
\usepackage{mathtools}
\usepackage{amsfonts}
\usepackage{gensymb}
\makeglossaries
\ifCLASSINFOpdf
   \usepackage[pdftex]{graphicx}
  \graphicspath{ {./images/} }

\else
\fi
\begin{document}
%
\title{ArUco Maker based localization and Node graph approach to mapping}
%
%
%

\author{Abhijith~Sampathkrishna
\thanks{A.Sampathkrishna is a student at the University of Glasgow and former student and research assistant of KLE Technological University}
}

\maketitle
\thispagestyle{empty}

\begin{abstract}
This paper explores a method of localization and navigation of indoor
mobile robots using a node graph of landmarks that are based on fiducial markers. The use of ArUco markers and their 2-D orientation with respect to the
camera of the robot and the distance to the markers from the camera is used to calculate the
relative position of the robot as well as the relative positions of other markers.
The proposed method combines aspects of beacon-based navigation and
Simultaneous Localization and Mapping (SLAM) based navigation. The implementation of this method uses a depth camera to obtain the distance to the marker. After calculating the required orientation of the marker, it relies on odometry calculations for tracking the robot’s position after localization with respect to the marker. Using the odometry and the relative position of one marker, the robot is then localized with respect to another marker. The relative positions and orientation of the two markers are then calculated. The markers are represented as nodes and the relative distances and orientations are represented as edges connecting the nodes and a node graph can be generated that represents a map for the robot. The method was tested on a wheeled humanoid robot with the objective of having it autonomously navigate to a charging station inside a room. This objective was successfully achieved and the limitations and future improvements are briefly discussed.

\end{abstract}

\begin{IEEEkeywords}
ArUco, Node graph, Mapping, Locomotion, Localization, Navigation.
\end{IEEEkeywords}

%
\IEEEpeerreviewmaketitle

\section{Introduction}
%
%
%
%
\IEEEPARstart{L}ocalization, Mapping and autonomous navigation of robots have been the subject of a great deal of research and development over the recent decades. Various approaches and methods have been developed for indoor autonomous navigation, many of them are primarily based on vision \cite{litreview}. Apart from vision based systems, various different approaches also exist that use beacons \cite{yi_beacon}\cite{blebeacon} and Simultaneous Localization and Mapping (SLAM) algorithms \cite{orbslam}\cite{monoslam}, the latter is perhaps the most popular method used for indoor navigation.
Beacon based systems, by definition, require one or multiple external devices placed in the environment; The robot uses the characteristics of the signals emitted by these devices such as strength and time-of-flight, to calculate the relative position of the robot within the environment. SLAM, however, does not require any external devices and works by trying to recognise the environment by extracting static features and tracking the movement of the robot, while trying to build a map of the environment which is commonly represented by an Occupancy Grid\cite{occgrid}.
Both of these systems have their own advantages and disadvantages; however, the aim of the proposed method in this paper attempts to find the optimal trade-off between adding extrinsic devices to the environment, as done in the beacon systems, and creating a detailed global map of the environment, as done in SLAM, and explore the effectiveness of navigating autonomously without having to localize the robot continuously, which is done in both of these methods, but creating intrinsic headings that point toward different landmarks. 

A majority of mobile robots that are being developed have cameras in the design and many of them use depth or stereo cameras. Taking advantage of this, a vision based localization and navigation system could be developed that uses the detection and recognition of the 2-D fiducial markers such as ArUco\cite{arucodetection} or AprilTags \cite{apriltag}. The encoding of information and the monochromatic design of these markers make the detection and localization of the markers within a frame quite fast and extremely efficient. It also uses much less processing power.  Visual localization using fiducial markers has been used in UAV navigation and landing algorithms \cite{uavapriltag}\cite{uavlocalization} and also for localizing a mobile robot \cite{vislocalization}. There has also been research that explores landmark based navigation that focuses on extracting features from the environment and recognising static environmental features \cite{landmarknav}.

\section{ArUco markers}
ArUco markers are made up of square grids of cells that are filled in with either black or white colours. The detection algorithms of such markers are highly reliable and accurate\cite{arucodetection}.  
\begin{figure}[h]
    \centering
    \includegraphics[scale=0.20]{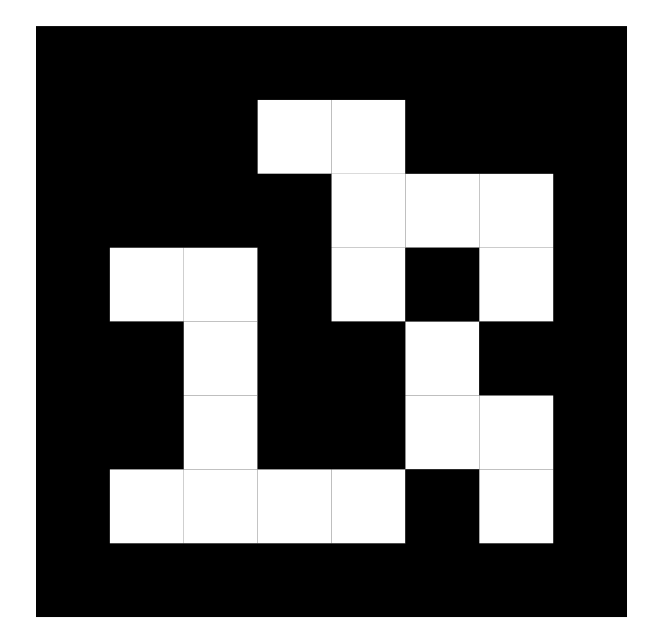}
    \caption{Example of an ArUco Marker}
    \label{fig:example_aruco}
\end{figure}
\begin{figure}[h]
    \centering
    \includegraphics[scale=0.60]{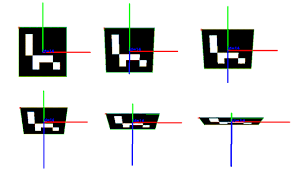}
    \caption{Detection in various orientations}
    \label{fig:example_aruco_ori}
\end{figure}

To generate and detect these markers, a minimal library has been created based on OpenCV \cite{speduparucodetection}\cite{arucogeneration}. The detection algorithms are developed to be able to detect the marker even when it is in different orientations and sizes (Fig.\ref{fig:example_aruco_ori}). This is the main feature that will be useful in relative localization.
This aspect of detecting the marker and the four corners makes extracting the orientation of the marker relative to the camera simple and efficient.

Using a single camera, it is possible to calculate the width, height and orientation of the marker relative to the camera, but this requires accurate camera parameters and possibly calibration \cite{vislocalization}\cite{singlecameranav}. However, it is not necessary to calculate all of the above mentioned parameters for the proposed method. Calculating the yaw angle of the marker will be sufficient.

\section{Localization}
Focusing on 2-D localization in an indoor environment, the localization method requires extracting two pieces of information from the marker; Relative Yaw angle between the normal vector of the marker plane and the line-of-sight of the camera, and the distance from the camera, these can then be used to create a two-coordinate system. There is a primary assumption in this setup:
\begin{itemize}
     \item The marker plane's normal vector and the camera plane's normal vector (line-of-sight) lie on the same plane or parallel planes. I.e, the camera plane and the marker plane (wall) are both perpendicular to the \(xy\) plane (floor).
\end{itemize}

The localization module would work primarily by localizing the robot relative to only one marker in polar coordinates. A global node graph can be created to link one marker to another. One of the markers will be defined as the "home marker" which will have the local and global coordinates of (0,0).

\begin{figure}[h]
    \centering
    \includegraphics[scale=0.32]{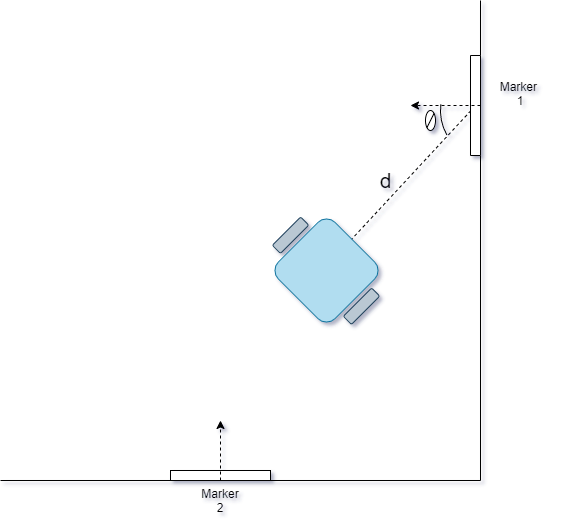}
    \caption{Illustration of the robot in the 2-D environment}
    \label{fig:illustration_1}
\end{figure}

From the figure (Fig.\ref{fig:illustration_1}), it can be observed that it is possible to establish a 2-D pose of the robot with respect to the marker with only two values; The angle \(\theta\) which is the rotation in Z-axis of the marker with respect to the camera, or Yaw, and the distance \(d\). 
Extracting the distance can be done in various ways depending on the type of camera in use. In experimentation and testing, a depth camera was used to obtain the \(d\) value. Calculating the \(\theta\) value, however, could be done using any camera without calibration.

\begin{figure}[h]
    \centering
    \includegraphics[scale=0.75]{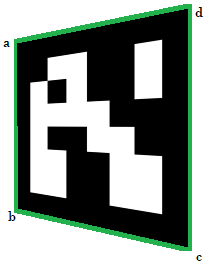}
    \caption{Marker rotated in Z-axis}
    \label{fig:yawillustration}
\end{figure}

Although the ArUco library in OpenCV provides a way of getting the orientation using Rodrigues rotation\cite{rodrigues}, it requires a calibrated set of camera parameters and distortion coefficients. However, to obtain the Yaw value, there is a simpler method.
The library returns a list of four points with pixel coordinates that contain the four corners of the marker; \((x_a,y_a),(x_b,y_b),(x_c,y_c)(x_d,y_d)\). 
Figure (Fig.\ref{fig:yawillustration}) represents a close approximation of what the robot might see according to the previous illustration of the environment setup (Fig.\ref{fig:illustration_1}). To calculate the Yaw \(\theta_z\); Firstly, the apparent horizontal side length \(s_a\) is calculated.

\[s_a = \frac{|x_a-x_d|+|x_c-x_b|}{2}\]

It is then be assumed that the ideal,or true, horizontal side length \(s_i\) to be equal to either one of the apparent vertical sides.
 \[s_i= |y_a-y_b|\] or \[s_i=|y_d-y_c|\]

And the value of \(\theta_z\) can be calculated by,
\[\theta_z = \arccos(\frac{s_a}{s_i})\]

Considering \(s_i\) to be the side \(ab\), \(\theta\) would be,
\[
    \theta= 
\begin{dcases}
    +\theta_z,& \text{if } |y_a-y_b|<|y_d-y_c|\\
    -\theta_z,& \text{if }|y_a-y_b|>|y_d-y_c|
\end{dcases}
\]

\begin{figure}[h]
    \centering
    \includegraphics[scale=0.30]{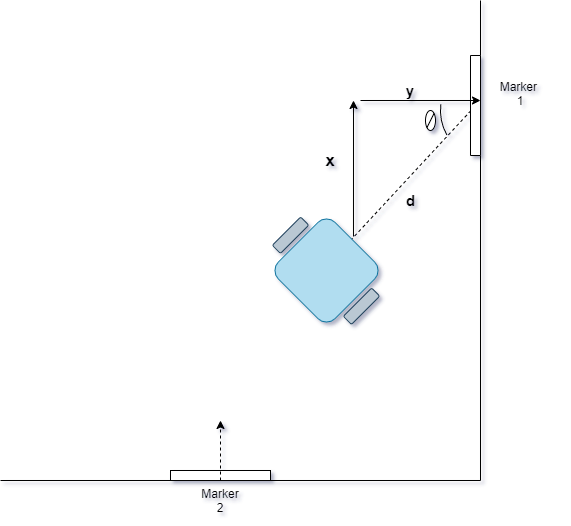}
    \caption{Implementation Trajectory}
    \label{fig:illustration_2}
\end{figure}
The implementation of this gave consistent and accurate measurements of \(\theta\) with an error of less than \(2\degree\) from a distance of \(3m\) or less with the marker size being \(20cm\times20cm\). It was also noted that the measurement is less accurate when the marker is not aligned at the center of the frame. The same issue was observed in regards to the distance measurement with the depth camera as well. Therefore, there was a need to align the camera to the center of the frame to obtain reliable results. Which actually provided an advantage when building the node graph.

\begin{figure}[h]
    \centering
    \includegraphics[scale=0.75]{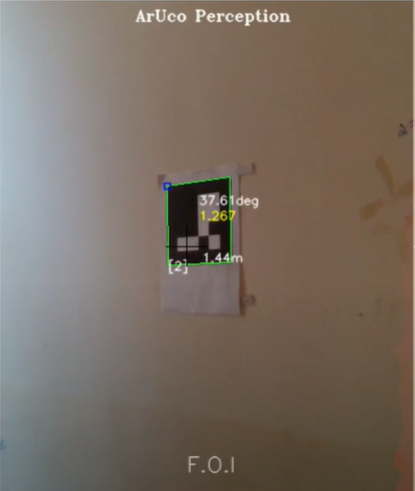}
    \caption{Experimental result of marker detection, angle $\theta=37.61deg$, distance (1.44m) and corrected distance to marker (1.267m) with Marker ID=2}
    \label{fig:marker_result}
\end{figure}

During implementation and testing, the Cartesian coordinates were also calculated using the standard trigonometric formulas to create a taxi-cab trajectory for the robot to go toward the marker. The values of \(x\) and \(y\), illustrated in figure (Fig.\ref{fig:illustration_2}), were calculated and the error was observed to be approximately less than or equal to \(15cm\) for both \(x\) and \(y\) from a distance \(d\) of up to \(4m\).

\section{Locomotion}

The implementation adopted a simple locomotion model of a rotate-translate-rotate algorithm. From any given position, the robot will consider any given goal in polar coordinates; It will first rotate to minimize the angular difference (\(\theta_{diff}=\theta_{goal}-\theta_{current}\)), and then translate to minimize the difference in distance (\(d_{diff}=d_{goal}-d_{current}\)). However, the current pose values \(\theta_{current}\) and \(d_{current}\) will be set to zero at the start of a new goal and the sensor data is offset and corrected for the same, this makes for a simple and effective method for instantaneous robot locomotion. 

Although the locomotion model works for instantaneous coordinate goals with the robot's frame of reference, there is a need to keep track of the robot based on historic motion. This can be achieved by calculating the odometric position of the robot with respect to an origin (0,0), which could be the coordinates of the home marker. 

\begin{figure}[h]
    \centering
    \includegraphics[scale=0.35]{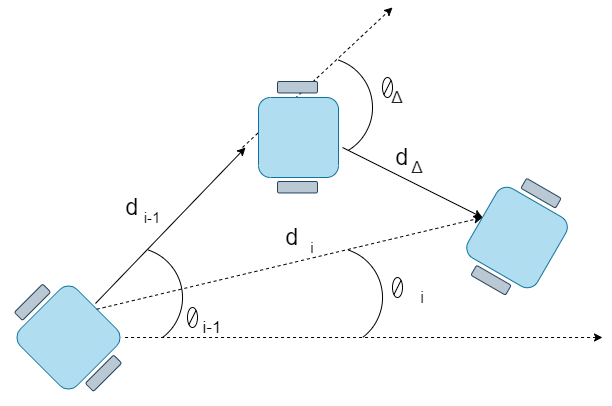}
    \caption{Odometry Illustration}
    \label{fig:odeometry_illustration}
\end{figure}

Consider the robot to be at \((0,0)\), with an initial goal of \((d_{i-1},\theta_{i-1})\); At instance \(i=1\) (the first goal) , the odometry \(O_i\) will simply be the initial goal. When the next goal is given to the robot, suppose \((d_\Delta,\theta_\Delta)\), the odometry \(O_i\), or \((d_i,\theta_i)\), can be calculated as follows,

\[
    d_i = \sqrt{d_{i-1}^2 + d_\Delta^2 - 2d_{i-1}d_\Delta\cos(\pi-\theta_\Delta)}\]
\[
    \theta_i = \arccos(\frac{d_i^2+d_{i-1}^2-d_\Delta^2}{2d_id_{i-1}})
\]

Every goal thereafter will be considered as \((d_\Delta,\theta_\Delta)\), or \(O_\Delta\); And with the previously calculated odometry \(O_{i-1}\), the new or current odometry can be expressed a function, say \(\alpha\), of \(O_{i-1}\) and \(O_\Delta\). 
\[ O_i = \alpha(O_{i-1},O_\Delta)\]

The application of this was tested on a humanoid robot for the purpose of autonomous navigation to a charging station, the trajectory illustrated in figure (Fig.\ref{fig:illustration_2}) was used to achieve this.

\begin{figure}[h]
    \centering
    \includegraphics[scale=0.13]{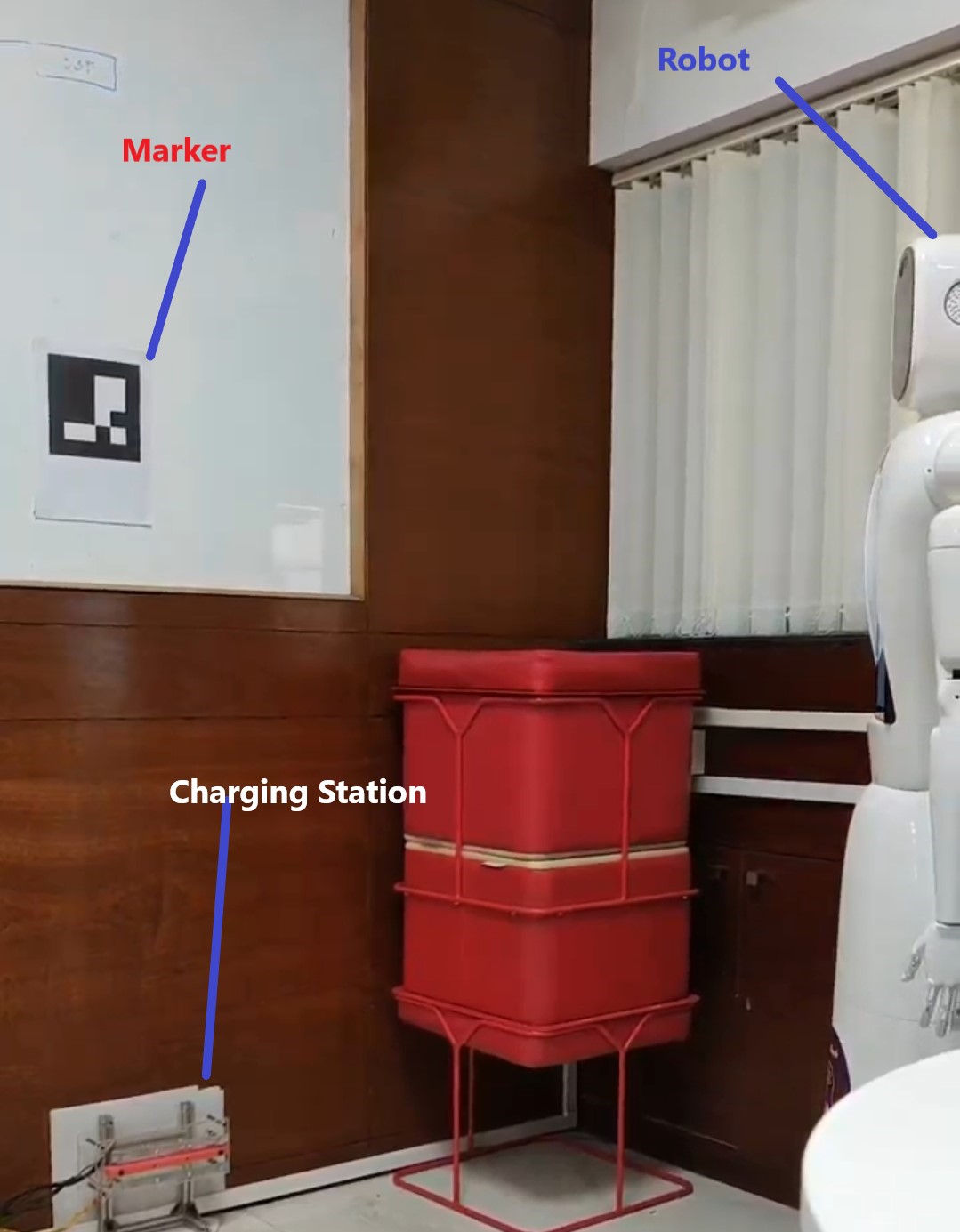}
    \caption{Initial Search for the marker}
    \label{fig:p1}
\end{figure}

\begin{figure}[h]
    \centering
    \includegraphics[scale=0.15]{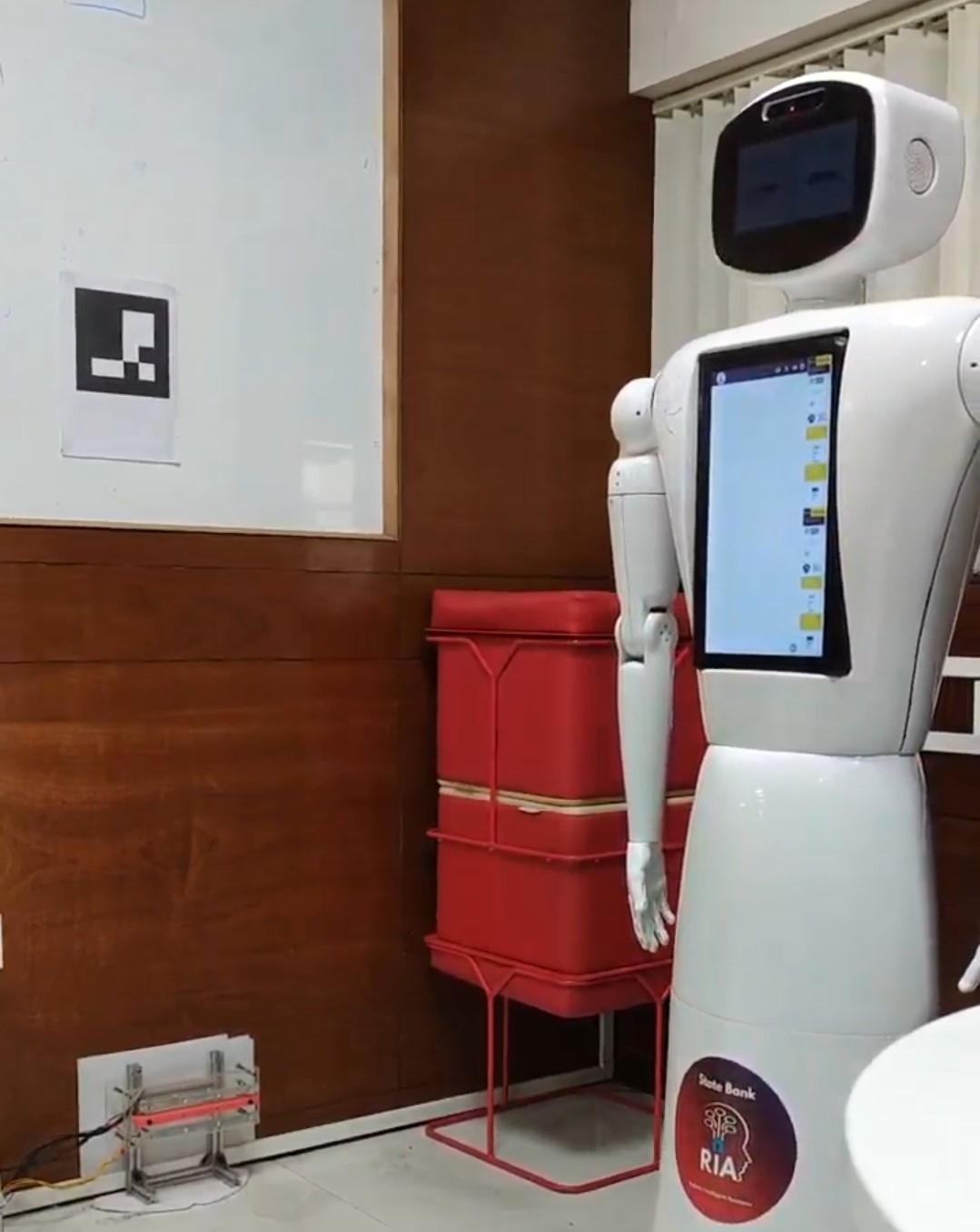}
    \caption{Move to first point}
    \label{fig:p2}
\end{figure}

\begin{figure}[h]
    \centering
    \includegraphics[scale=0.15]{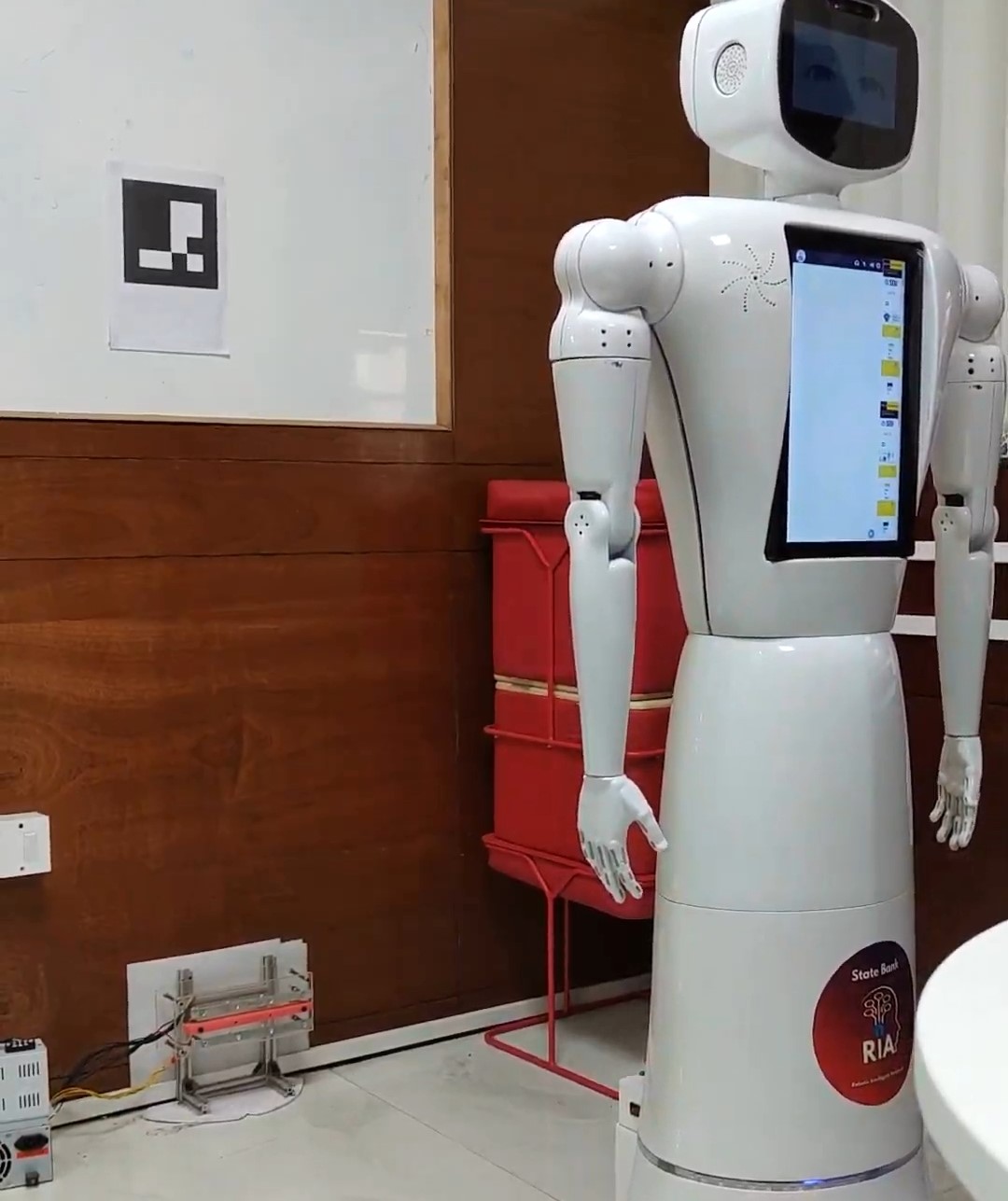}
    \caption{Rotate to align charging probes towards the charging station}
    \label{fig:p3}
\end{figure}

\begin{figure}[h]
    \centering
    \includegraphics[scale=0.15]{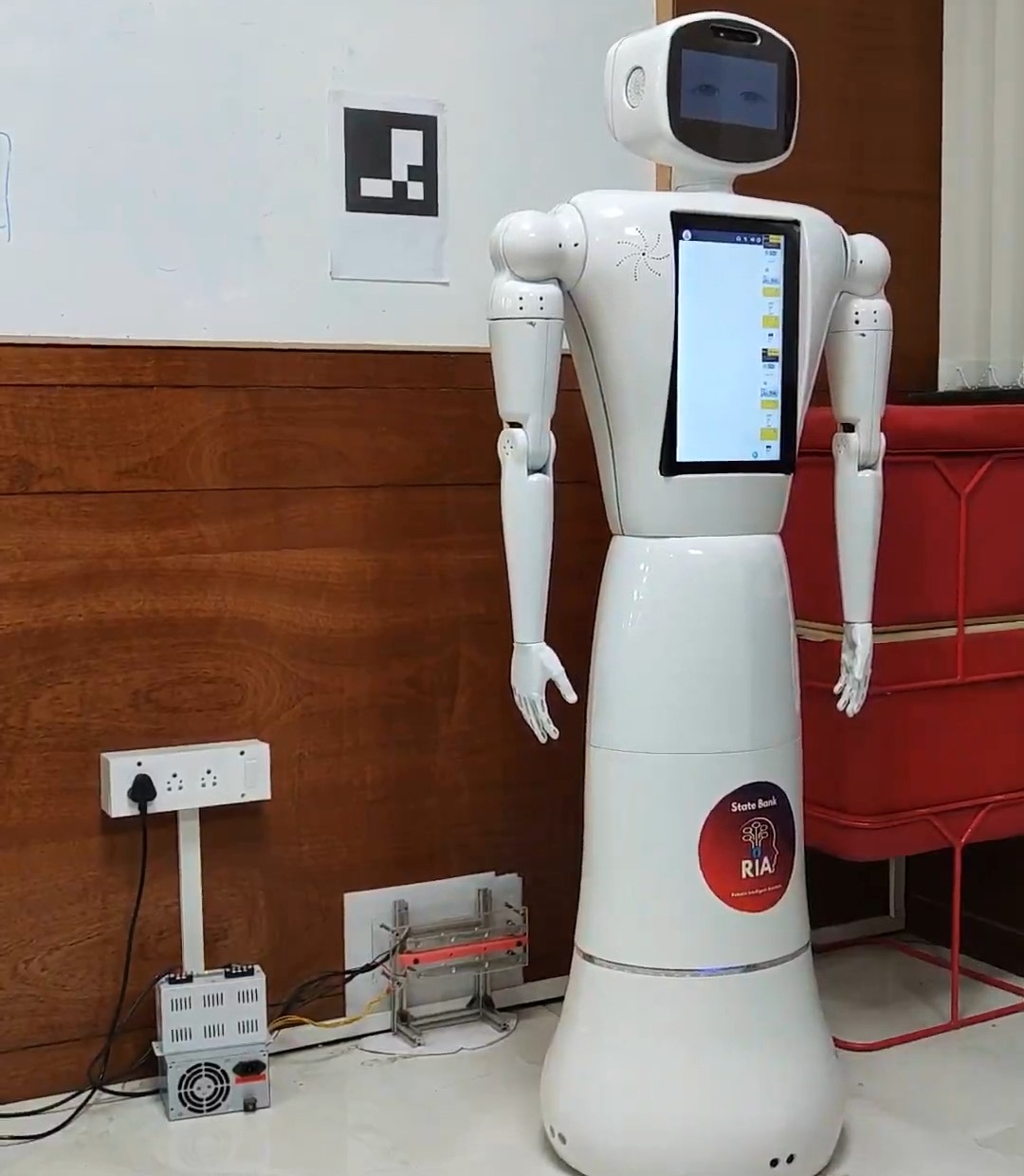}
    \caption{Move towards the charging station/marker}
    \label{fig:p4}
\end{figure}

Figures \ref{fig:p1},\ref{fig:p2},\ref{fig:p3} and \ref{fig:p4} show the trajectory executed by the robot autonomously after it found the marker that was aligned with the charging station.

\section{Mapping}

This method proposes generating a network of nodes, or a graph, similar to graph based SLAM\cite{graphslam}; using the markers as nodes, the straight-line distance and three angular differences between them. A similar method to odometry calculation can be used to find the phase difference \(\phi_{ab}\) , the angular differences \(\theta_{ab}\) and \(\theta_{ba}\) with respect to their normal vectors and and their straight-line distance \(d_{ab}\) between markers \(a\) and \(b\). These values can be used to establish a connection between the two markers in 2-D space.

\begin{figure}[h]
    \centering
    \includegraphics[scale=0.50]{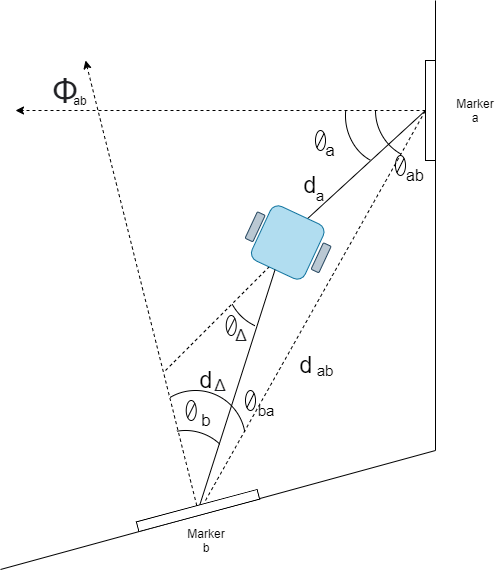}
    \caption{Linking Two markers in 2-D space}
    \label{fig:twomarker}
\end{figure}

In the figure (Fig.\ref{fig:twomarker}), the direction \(\theta_a\) and distance \(d_a\) between the robot and marker \(a\) could correspond to an odometry at that instant, say \(O_a\), which would have been calculated from the time the robot was localized with respect to marker \(a\). The angle \(\theta_\Delta\) would be the angle it takes the robot to align itself with marker \(b\). Then the values for \(\theta_b\) and \(d_\Delta\) (or \(d_b\)) can be calculated using the methods explained in the previous sections.

For building a comprehensive map that the robot could use to calculate it's relative position with respect to any marker, given that it has already localized itself with one marker; the phase difference \(\phi_{ab}\), the distance \(d_{ab}\), and the angles \(\theta_{ab}\) and \(\theta_{ba}\) would be required. It is possible to calculate all of these values as follows,

\[    d_{ab} = \sqrt{d_a^2+d_\Delta^2-2d_ad_\Delta\cos(\pi-\theta_\Delta)} \]
\[    \theta_{ab} = \theta_a + \arccos(\frac{d_a^2+d_{ab}^2-d_\Delta^2}{2d_ad_{ab}}) \]
\[    \theta_{ba} = \theta_b + \arccos(\frac{d_\Delta^2+d_{ab}^2-d_a^2}{2d_\Delta d_{ab}})\]
\[    \phi_{ab} = \pi - (\theta_{ab}+\theta_{ba})    \]

The spatial connection, say \(E\), between any two given markers \(a\) and \(b\) can now be expressed as a set,
\[ E_{ab} = \{\phi_{ab},\theta_{ab},\theta_{ba},d_{ab}\}\]

Given \(E_{ab}\) and \(E_{bc}\), it is then possible to calculate \(E_{ac}\).

\begin{figure}[h]
    \centering
    \includegraphics[scale=0.45]{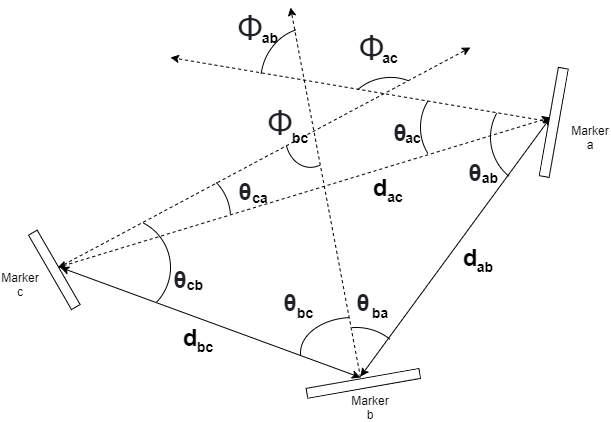}
    \caption{Relationship between three markers}
    \label{fig:threemarkers}
\end{figure}

From the figure (Fig.\ref{fig:threemarkers}), Assuming \(E_{ab}\) and \(E_{bc}\) are already computed, \(E_{ac}\) can be calculated as below, with the known values:
\[ E_{ab} = \{\phi_{ab},\theta_{ab},\theta_{ba},d_{ab}\}\]
\[ E_{bc} = \{\phi_{bc},\theta_{bc},\theta_{cb},d_{bc}\}\]

Then,
\[ \phi_{ac} = \phi_{ab}+\phi_{bc} \]
\[ d_{ac} = \sqrt{d_{ab}^2+d_{bc}^2-2d_{ab}d_{bc}\cos(\theta_{ba}+\theta_{bc})}\]
\[ \theta_{ac} = \theta_{ab}- \arccos(\frac{d_{ab}^2+d_{ac}^2-d_{bc}^2}{2d_{ab}d_{ac}})\]
\[ \theta_{ca} =  \pi-\phi_{ac}+\theta_{ac} \]
\[ E_{ac} = \{\phi_{ac},\theta_{ac},\theta_{ca},d_{ac}\}\]
This can be represented as a function \(M\),
\[E_{ac} = M(E_{ab},E_{bc})\]

A graph could now be constructed considering each marker as a node with it's unique ArUco ID, and an edge between two nodes can be represented by \(E\).

\begin{figure}[h]
    \centering
    \includegraphics[scale=0.25]{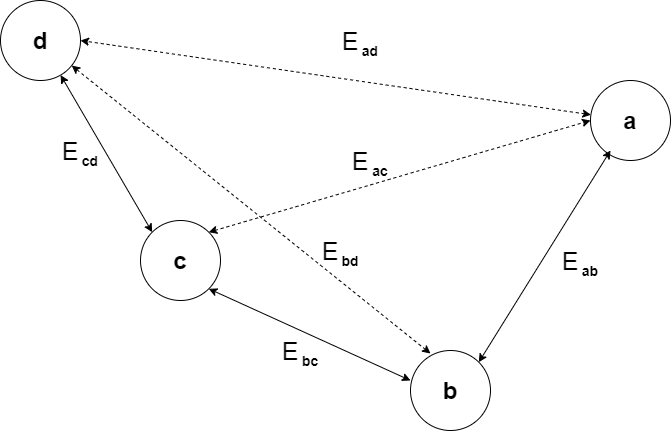}
    \caption{Graph of marker nodes}
    \label{fig:graph}
\end{figure}

In the above sample graph (Fig.\ref{fig:graph}), \(E_{ab}\),\(E_{bc}\) and \(E_{cd}\) are obtained through the robot. \(E_{ac}\),\(E_{bd}\) and \(E_{ad}\) can be calculated with the help of \(M\),
\[E_{ac} = M(E_{ab},E_{bc})\]
\[E_{bd} = M(E_{bc},E_{cd})\]
\[E_{ad} = M(E_{ac},E_{cd})\]

\section{Target Heading Calculation}

Considering the robot has localized with respect to a marker, say \(a\), and needs to go toward a target marker, \(c\). A goal vector, or heading, can be generated based on \(E_{ct}\).

\begin{figure}[h]
    \centering
    \includegraphics[scale=0.30]{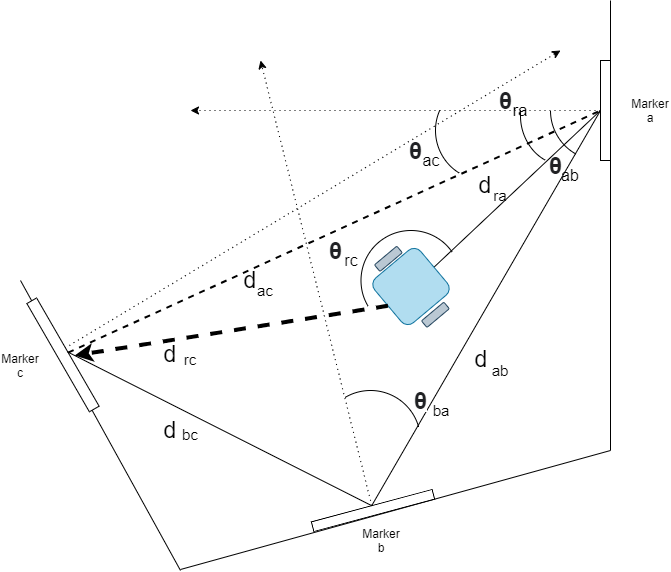}
    \caption{Heading Calculation}
    \label{fig:heading}
\end{figure}

In the figure (Fig.\ref{fig:heading}), the target marker is \(c\) and the localized marker is \(a\). In order to calculate the desired goal \(O_{heading}\), \(E_{ac}\) is first calculated, if it is not calculated intrinsically with the robot, with the help of the function \(M\).
\[E_{ac} = M(E_{ab},E_{bc})\]
Then with the help of the values from \(E_{ac}\), the goal heading can now be calculated as,

\[d_{heading} = \sqrt{d_{ac}^2+d_{ra}^2-2d_{ra}d_{ac}\cos(\theta_{ac}-\theta_{ra})}\]
\[\theta_{heading} = \arccos(\frac{d_{ra}^2+d_{heading}^2-d_{ac}^2}{2d_{ra}d_{heading}})\]
\[O_{heading} = (d_{heading},\theta_{heading})\]

The robot can now keep track of the difference in direction, \(\theta_{diff}=(\theta_{heading}-\theta_{current})\), and go towards the target marker with the navigation model.

\section{Conclusion}
This paper has proposed a method to localize and navigate a mobile robot in an indoor environment, and create a graph based map of the environment based on static ArUco markers, considering them as nodes. It explored the required information that needs to be extracted from the marker in order to localize the robot in a 2-D space. A simple navigation model was also discussed which compliments and helps the localization and mapping. The procedure and the underlying mathematics to build a graph with each marker as a node was explored. The implementation of this method was carried out with a depth camera, Intel Realsense D435i, to obtain the distance information. However, the alternative method proposed in the paper to extract Yaw values was tested using different cameras with consistent results. A humanoid robot that was developed in-house at KLE Technological University was used to test the locomotion and mapping aspects of the proposed method. During implementation, a map of three markers were tested successfully within a single room. Accumulated drift causing the robot to stray after travelling continuously between the markers were observed and deemed to be significant without using any filters or corrective algorithms. 

\section{Future Scope}
During further development, noise filtering algorithms, such as the Extended Kalman Filter, could be used in order to minimize drift and attain a more accurate system. Trajectory planning to avoid obstacles and navigate to different nodes needs to be explored and developed. It would be useful to find an alternative to markers that make it possible to obtain information similar to the proposed method with ArUco markers that can be used to create a two-coordinate system without using artificial markers to avoid having to modify the environment at any scale. 


%

\section*{Acknowledgment}

The authors would like to thank the teaching and non-teaching faculty of the Automation and Robotics department of KLE Technological University for their support and encouragement throughout the project. They would also like to thank Alvin M Reji and Prithvi Deshpande for their efforts and assistance in the project. 

\ifCLASSOPTIONcaptionsoff
  \newpage
\fi



%
\bibliographystyle{unsrt}
\bibliography{references}
%




\end{document}